\documentclass[runningheads]{llncs}

\usepackage[utf8]{inputenc}
\usepackage[T1]{fontenc}
\usepackage{graphicx}
\usepackage{textcomp}
\usepackage{microtype}
\usepackage{cite}

\usepackage{amsmath,amssymb,amsfonts}
\usepackage{mathtools}
\usepackage{bm}
\usepackage{nicefrac}

\usepackage{algorithm}
\usepackage{algorithmic}
\renewcommand{\algorithmicrequire}{\textbf{Input:}}
\renewcommand{\algorithmicensure}{\textbf{Output:}}

\usepackage{booktabs}
\usepackage{multicol}
\usepackage{multirow}
\usepackage{tabularx}
\usepackage{makecell}
\usepackage{threeparttable}
\usepackage{subcaption}   %
\usepackage{placeins}
\usepackage{float}

\usepackage{xspace}
\usepackage[normalem]{ulem}
\usepackage{paralist}
\usepackage{xcolor}
\usepackage{bbding}       %
\usepackage{url}
\usepackage{xurl}

\usepackage[colorlinks=true,linkcolor=black,citecolor=black,urlcolor=black]{hyperref}
\usepackage{orcidlink}
\usepackage[capitalize,noabbrev]{cleveref}

\newcommand{\ie}{\emph{i.e.,}\xspace}
\newcommand{\eg}{\emph{e.g.,}\xspace}
\newcommand{\iid}{\emph{i.i.d.}\xspace}

\newcommand{\sys}{LDPKiT\xspace}
\newcommand{\sysR}{LDPKiT-Rand\xspace}
\newcommand{\sysS}{LDPKiT-Sup\xspace}
\newcommand{\privdata}{{\mathcal{D}_{\rm priv}}}
\newcommand{\inferdata}{{\mathcal{D}_{\rm infer}}}
\newcommand{\protectdata}{{\mathcal{D}_{\rm protected}}}

\newcommand{\valdata}{{\mathcal{D}_{\rm val}}}

\newcommand{\teacher}{{\mathcal{M}_{\rm R}}}
\newcommand{\student}{{\mathcal{M}_{\rm L}}}

\newif\ifusetodo
\newif\ifshowcomments

\showcommentstrue
\usetodotrue

\ifshowcomments
\ifusetodo
    \usepackage[textsize=scriptsize]{todonotes}
    \newcommand{\david}[2][]{\todo[color=green!30,#1]{David: #2}}
    \newcommand{\kexin}[2][]{\todo[color=orange!30,#1]{Kexin: #2}}
    \newcommand{\aastha}[2][]{\todo[color=blue!30,#1]{Aastha: #2}}

\else
    \newcommand{\david}[1]{{\color{purple}(David: #1)}}
    \newcommand{\kexin}[1]{{\color{orange}(Kexin: #1)}}
    \newcommand{\aastha}[1]{{\color{blue}(Aastha: #1)}}
    \newcommand{\todo}[1]{{\color{blue} (TODO: #1)}}

\fi
\else
    \newcommand{\david}[2][]{\ignorespaces}
    \newcommand{\kexin}[2][]{\ignorespaces}
    \newcommand{\aastha}[2][]{\ignorespaces}

    \newcommand{\todo}[1]{\ignorespaces}

\fi
\usepackage{comment}
\spnewtheorem{assumption}[theorem]{Assumption}{\bfseries}{\itshape}
\spnewtheorem{hypothesis}[theorem]{Hypothesis}{\bfseries}{\itshape}

\begin{document}

\title{LDPKiT: Superimposing Remote Queries for Privacy-Preserving Distillation}
\titlerunning{LDPKiT: Privacy-Preserving Distillation}

\author{
Kexin Li\inst{1}\Envelope\orcidlink{0000-0002-7911-3731}
\and
Aastha Mehta\inst{2}\orcidlink{0009-0005-3416-5254}
\and
David Lie\inst{1}\orcidlink{0000-0002-2000-6827}
}
\authorrunning{K. Li et al.}

\institute{
University of Toronto, Toronto, ON, Canada\\
\and
The University of British Columbia, Vancouver, BC, Canada\\
\email{cassiekx.li@mail.utoronto.ca}
}

\maketitle

\begin{abstract}
To protect privacy in regulated domains such as healthcare and finance, model owners may allow only remote API access while keeping both the training data and model parameters private. However, model users performing inference on such remotely hosted models may be required to transmit potentially sensitive inputs, raising privacy concerns.
In this work, we present \sys{}, a framework for non-adversarial, privacy-preserving model distillation that leverages a user's private in-distribution data while bounding privacy leakage. \sys{} introduces a novel superimposition technique that generates approximately in-distribution samples, enabling effective knowledge transfer under local differential privacy (LDP).
Experiments on Fashion-MNIST, SVHN, and PathMNIST demonstrate that \sys{} consistently improves utility while maintaining privacy, with benefits that become more pronounced at stronger noise levels. For example, on SVHN, \sys{} achieves nearly the same inference accuracy at $\epsilon=1.25$ as at $\epsilon=2.0$, yielding stronger privacy guarantees with less than a 2\% accuracy reduction. We further conduct sensitivity analyses to examine the effect of dataset size on performance and provide a systematic analysis of latent space representations, offering intuitive and empirical insights into the accuracy gains of \sys{}.

\end{abstract}

\section{Introduction}\label{sec:intro}

A machine learning (ML) model, trained on a dataset, is a representation of the data that can be used to make predictions and offer insights on other data that is of a similar distribution to the original training set. As a result, even in cases where the training data itself cannot be shared due to privacy restrictions, there is still a strong incentive to share access to the ML model. Moreover, techniques such as PATE~\cite{papernotPATE} and federated learning~\cite{zhao2021comprehensive} have been proposed to enable training such models while limiting the privacy impact on the training set. The privacy protection such techniques confer is maximized when offering only black-box access (\ie keeping the weights private) to such models, so that the intermediate activations computed during inference remain hidden~\cite{shokri2021privacyrisksmodelexplanations}.

Sensitive data is sometimes shared for socially beneficial reasons, particularly in regulated sectors where organizations must disclose privacy-sensitive information to protect public welfare. For example, hospitals may be required to share patient information during disease outbreaks, and financial institutions may need to report accounts suspected of money laundering. ML could enable similar benefits while reducing the need to share sensitive datasets directly: organizations with large proprietary datasets could train predictive models using privacy-protective techniques~\cite{papernotPATE, zhao2021comprehensive} and provide controlled access to entities without comparable data. For instance, a healthcare network could offer smaller clinics access to a model that predicts disease risk, while a large financial institution could provide local banks with a model for detecting money laundering.

However, while black-box models trained using privacy-preserving methods protect the training set, inference in the black-box setting still requires the model user to transmit potentially sensitive inputs.
While various methods for protecting privacy during inference have been proposed, none satisfy the requirements of efficiency and ease of adoption.
For instance, homomorphic encryption~\cite{gazelle} can impose severe overheads and typically targets an honest-but-curious adversary model, while hardware-enforced trusted execution environments~\cite{tramer2019slalom} require specialized hardware that is currently not broadly available.

The strongest privacy guarantee available to a model user is to obtain the trained model weights and perform inference locally, so that the user's data never leaves their control. However, this would open up the original model to white-box attacks.
To address this, we propose a privacy-preserving distillation of black-box models while bounding the privacy leakage of query inputs, which we call
\sys{} (\textit{Local Differentially-Private and Utility-Preserving Inference via Knowledge Transfer}).
Effective distillation requires users' queries to be representative of the model's training data distribution, so ideally, they should use real data. However, in regulated privacy-sensitive domains, users typically possess only small amounts of such data, \eg~smaller clinics may have far fewer patients. \sys{} addresses the privacy and data-shortage challenges by applying local differential privacy (LDP) to bound the privacy leakage of each query while introducing a data augmentation technique that superimposes pairs of noised inputs to generate additional in-distribution queries. This combination enables users to distill useful knowledge from remote models while ensuring protection of sensitive query data.

Our analysis is guided by the following research questions:

\textbf{RQ1.} Does \sys{} recover the utility impacted by LDP noise? (Section~\ref{subsec:RQ1eval})

\textbf{RQ2.} Why does superimposition work in \sys{}? (Section~\ref{subsec:latent})

\textbf{RQ3.} How do the size of the private dataset and the number of queries impact \sys{}? (Section~\ref{subsec:dataset_sensitivity})

\textbf{Contributions.}
\sys{} enables privacy-preserving distillation from remote black-box models by generating augmented inference datasets from users' $\epsilon$-LDP-protected private samples. Across three image-classification datasets and two models, \sys{} recovers much of the utility lost to LDP noise. We further study the impact of private-dataset size and number of queries, and conduct latent-space analysis to uncover factors contributing to \sys{}'s effectiveness.

\section{Overview}\label{sec:background}

\paragraph{\textbf{Motivation.}}
Our scenario considers privacy-sensitive ML services used to label privacy-sensitive data. 
Effective distillation for privacy requires realistic in-distribution data, which may itself be sensitive and unavailable publicly. Using out-of-distribution data is ineffective. For instance, distilling a ResNet-18 trained on Fashion-MNIST with SVHN samples, or vice versa, yields approximately random-guessing accuracy (\ie~10\%) for these 10-class datasets. By contrast, noisy in-distribution samples perform better: models distilled using $\epsilon$-LDP data with $\epsilon=2.0$ achieve roughly 21\% accuracy on SVHN and 28\% on Fashion-MNIST.

\paragraph{\textbf{Preliminaries and General Setup.}}
Our scenario involves two parties: a model provider hosting a remote model, $\teacher$, and a user with a small, unlabelled, sensitive dataset, $\privdata$, who distills a local model, $\student$. The user may be an organization or individual with black-box access to $\teacher$. To protect $\privdata$, the user applies $\epsilon$-LDP noise to each sample, producing $\protectdata$, where $|\privdata| = |\protectdata|$. Since $|\privdata|$ is typically small, we augment $\protectdata$ to form an expanded inference dataset, $\inferdata$. Its size, $|\inferdata|$, is determined by the user's query budget for $\teacher$, allowing the user to select some or all augmented samples. These samples are labelled by $\teacher$ and used to train $\student$.
We compare against \textit{SIDP}, a baseline in which the user directly queries $\teacher$ with $\protectdata$ and accepts the returned predictions, despite errors caused by LDP noise. 
To improve accuracy on $\privdata$, \sys{} distills $\student$ using samples from $\inferdata$ generated by two augmentation methods, \textit{\sysR} and \textit{\sysS}.

\paragraph{\textbf{Threat Model.}}
The user's goal is to label samples in $\privdata$ with reasonable accuracy when interacting with an ML service in a privacy-preserving way. We only assume that $\teacher$ provides faithful responses so that the user can obtain faithful predictions on inputs. Furthermore, the service provider has complete control of the model implementation and, as a result, has full access to the inference inputs. We note that the assumption of faithful responses is not required for \sys{}'s privacy guarantees, only for the utility of the $\student$.
To mitigate the threat of exposing sensitive data, the user applies $\epsilon$-LDP to their data before querying $\teacher$, limiting the amount of information that can be inferred from individual samples.
Our privacy goal is to protect the user's inference inputs from the remote model provider; \sys{} does not aim to protect the provider's proprietary model beyond the provider's existing black-box access policy.

\paragraph{\textbf{Privacy Guarantee.}}
\sys{} ensures the privacy of $\privdata$ during inference by applying LDP noise to each data sample. 
Intuitively, $\epsilon$-LDP ensures that after observing a protected query, the provider cannot confidently distinguish which raw private input produced it; a smaller $\epsilon$ makes inputs harder to distinguish.

\begin{definition} 
\textbf{${\epsilon}$-Local Differential Privacy (LDP).}\label{def:ldp}
We define ${\epsilon}$-LDP  as follows~\cite{dwork2006calibrating}:
A randomized algorithm $\mathcal{A}$ satisfies ${\epsilon}$-LDP if for all pairs of values and all sets $\mathcal{S}$ of possible outputs, where $\mathcal{S} \subseteq \text{Range}(\mathcal{A})$, 
\begin{equation}\label{eq:dp}
    \Pr[\mathcal{A}(v_1) \in \mathcal{S}] \leq e^{\epsilon} \Pr[\mathcal{A}(v_2) \in \mathcal{S}]
\end{equation}
\end{definition}
A lower $\epsilon$ indicates a tighter bound in the inequality and a stronger privacy guarantee.
We choose $\epsilon$ values primarily based on common industry standards~\cite{apple_white, Orr_2017} and empirically verify that protected data is resilient to data reconstruction attacks~\cite{schwethelm2025visualprivacyauditingdiffusion}.
We apply Laplace noise to satisfy the $\epsilon$-LDP privacy guarantee.

\begin{definition} \label{def:lap}

\textbf{Laplace Mechanism.}
For $x \in \mathcal{X}$, let
$\Delta_1 = \max_{x,x' \in \mathcal{X}} \|x-x'\|_1$ denote the $\ell_1$ sensitivity under the chosen neighboring relation. The mechanism releases $\tilde{x} = x + \eta$, where each coordinate $\eta_k \sim \operatorname{Lap}(0,\lambda)$ independently and $\lambda = \Delta_1/\epsilon$. This calibration satisfies $\epsilon$-LDP for each protected sample. For a scalar variable $t$, the probability density function of the Laplace distribution is
\begin{equation}
p_{\eta}(t; \lambda) = \frac{1}{2\lambda} \exp\left(-\frac{|t|}{\lambda}\right).
\end{equation}
\end{definition}

For constructing the inference dataset $\inferdata$, we use LDP's post-processing property: any processing of an $\epsilon$-LDP output remains $\epsilon$-LDP~\cite{dwork_eps_delta}. This allows LDPKiT to improve utility through augmentation without weakening privacy~\cite{Wang2021PostprocessingDP}.

\begin{definition} \label{def:post}
\textbf{Post-Processing Property in ${\epsilon}$-LDP.}
The post-processing property of LDP~\cite{dwork_eps_delta} states that if a randomized algorithm $\mathcal{A}$ satisfies ${\epsilon}$-LDP, then for any deterministic or randomized function $g$ independent from the original input values, the composed mechanism $g(\mathcal{A}(\cdot))$ also satisfies ${\epsilon}$-LDP. Specifically, for all $v_1, v_2$ (\ie~all values in the domain) and for all subsets $\mathcal{T} \subseteq \text{Range}(g(\mathcal{A}))$:
\begin{equation} \label{eq:postprocess}
    \Pr[g(\mathcal{A}(v_1)) \in \mathcal{T}] \leq e^{\epsilon} \Pr[g(\mathcal{A}(v_2)) \in \mathcal{T}].
\end{equation}

\end{definition}

\begin{figure}[t!]
    \centering
    \includegraphics[width=0.85\linewidth]{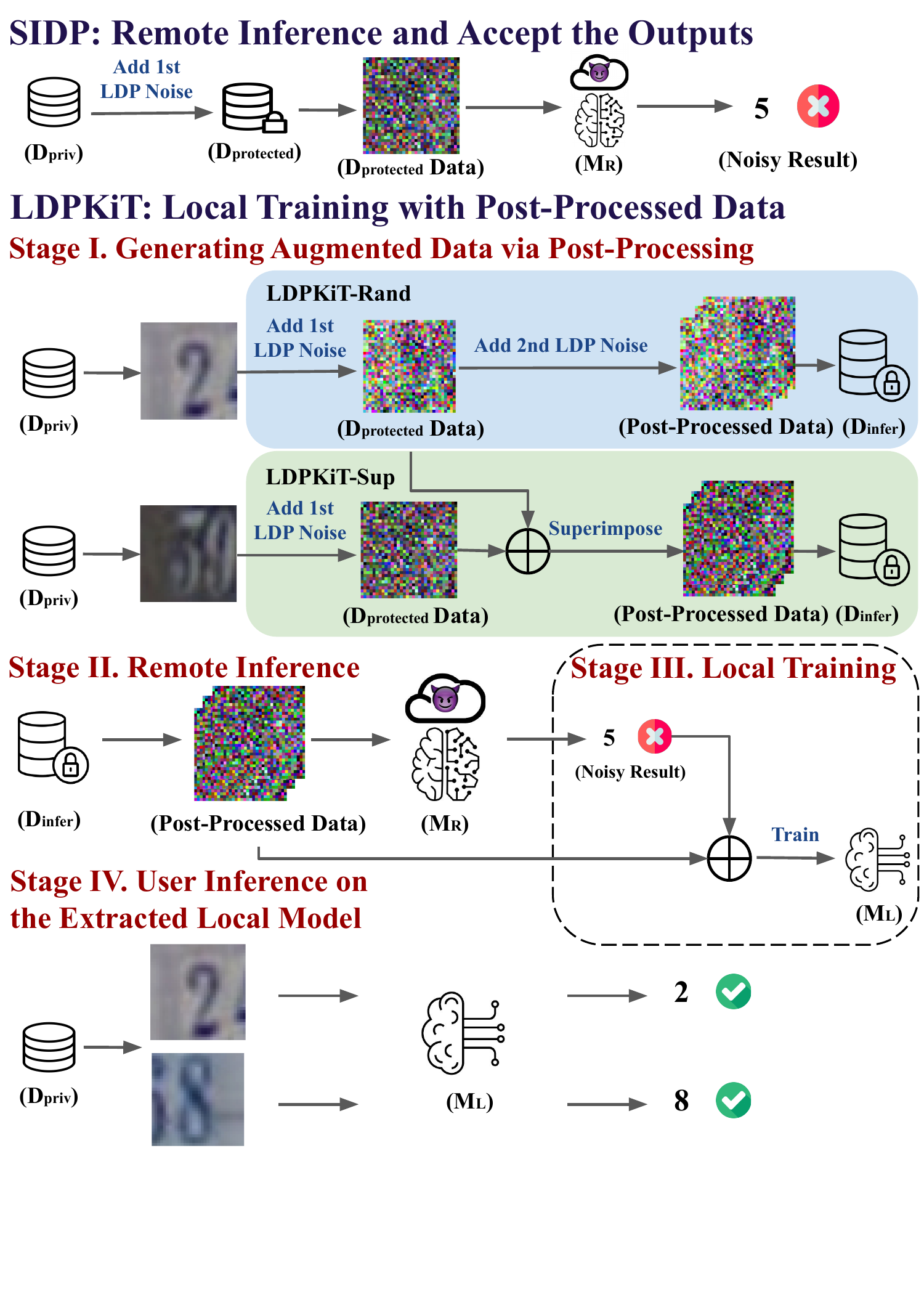}
    \caption{\sys system overview.}
    \label{fig:system_overview} 
\end{figure} 
\section{Design}\label{sec:overview}

Figure~\ref{fig:system_overview} summarizes SIDP and \sys{}. Since $|\protectdata| < |\inferdata|$, SIDP has a lower query cost than \sys{}, but at the cost of utility. SIDP directly queries $\teacher$ with $\protectdata$ and accepts the returned labels, while \sys{} post-processes $\protectdata$ into a larger $\inferdata$, queries $\teacher$ for labels, and trains $\student$ for local inference on $\privdata$. Thus, SIDP has lower query cost, whereas \sys{} uses additional post-processed queries to recover utility.

\paragraph{\textbf{Preliminary Experiments.}}
Distillation with OOD public data is ineffective (Section~\ref{sec:background}); hence, users must query $\teacher$ with in-distribution samples from $\privdata$. To protect these samples, \sys{} applies an $\epsilon$-LDP mechanism to obtain $\protectdata$. However, $\privdata$ is often small, and LDP noise degrades utility. \sys{} therefore augments $\protectdata$ into a larger inference dataset, $\inferdata$, while preserving privacy through LDP's post-processing property.
$\protectdata$ retains sufficient distributional information for distillation, yielding 2--3$\times$ higher utility than OOD data (Section~\ref{sec:background}). A natural augmentation strategy is to generate $n$ perturbed variants of each protected sample, producing $|\inferdata| = n \times |\protectdata|$. Since augmentation operates only on $\protectdata$, the $\epsilon$-LDP guarantee is preserved.

We validate this idea on SVHN using 500 randomly selected private samples, ResNet-152 as $\teacher$, and ResNet-18 as $\student$. Results are averaged over three runs and three dataset splits. With $\epsilon=1.5$, SIDP achieves only 14.16\% inference accuracy, indicating that $\teacher$ provides limited but nontrivial information under privacy protection. We then add a layer of random noise 499 times to each protected sample, yielding $|\inferdata|=249{,}500$ augmented queries. Distillation on $\inferdata$ improves $\student$ accuracy to 34.07\%, whereas distillation using only the 500 samples in $\protectdata$ yields near-random accuracy (\ie 10\%). Compared with SIDP, this gives a 20\% absolute accuracy improvement without additional privacy loss; a dependent two-sample t-test confirms statistical significance ($p<0.05$).

Nevertheless, this na\"ive random-noise augmentation is limited by the small size of $\privdata$ and the undirected nature of the perturbations. We hypothesize that more informative queries lie near decision boundaries in the latent data distribution. Motivated by this, we explore a superimposition-based augmentation strategy that refines post-processed LDP samples and better captures decision boundaries. Under the same setup, this approach improves ResNet-18 accuracy to 45.69\%, a further 10\% gain over random-noise augmentation.

\subsection{\sys{}'s Data Augmentation Mechanism}\label{subsec:noiseinjection}

\sys{} first applies the $\epsilon$-LDP mechanism~\cite{dwork2006calibrating} to each private sample in $\privdata$ once, producing $\protectdata$. All subsequent augmentation operates only on $\protectdata$ and data-independent randomness. We consider two post-processing strategies: \sysR{}, which adds additional random perturbations to protected samples, and \sysS {}, which averages pairs of protected samples.

\paragraph{\textbf{Mechanism-level Privacy Guarantee.}}
Let $\mathcal{M}(\privdata)$ denote the complete user-side procedure. 
Each private sample $d_{\text{priv}_i} \in \privdata$ is randomized once by an $\epsilon$-LDP
mechanism $\mathcal{A}_{\epsilon}$ to produce $d_{\text{protected}_i} = \mathcal{A}_{\epsilon}(d_{\text{priv}_i})$, using the Laplace mechanism in Definition~\ref{def:lap}. 
The augmented query set $\inferdata$ is then constructed only from $\{d_{\text{protected}_i}\}$ using data-independent post-processing, including additional random perturbation (\sysR{}) or pairwise averaging (\sysS{}).

For any two private datasets $D$ and $D'$ that differ in one record, all protected records except the changed one have the same distribution, while the changed record is protected by $\mathcal{A}_{\epsilon}$. Since the construction of $\inferdata$ and the selection of any query set are post-processing of protected records, for any event $T$ over query sets, $\Pr[\mathcal{M}(D) \in T] \leq e^{\epsilon} \Pr[\mathcal{M}(D') \in T]$. Thus, the query set $\inferdata$ preserves record-level $\epsilon$-LDP, assuming each private record is randomized once and later augmentations access only protected records and data-independent randomness.

\textbf{\sysR{}.} 
We formally describe the augmentation technique: for each $d_{\text{priv}_i} \in \privdata$, we construct an augmented dataset of the form: $d_{\text{infer}_i} \in \inferdata = d_{\text{priv}_i} + \mathcal{L}_{i} + \mathcal{L}_{j}$, where $\mathcal{L}_{i}$ and $\mathcal{L}_{j}$ are Laplace noise samples drawn independently with the same scale. The first noise layer, $\mathcal{L}_{i}$, is used to generate the initial privacy-protected dataset: $d_{\text{protected}_i} = d_{\text{priv}_i} + \mathcal{L}_{i}$, where $i = 1, 2, \dots, |\protectdata|$. We then apply the second noise layer, $\mathcal{L}_{j}$, where $j = 1, 2, \dots, |\protectdata| - 1$, as post-processing to create an expanded augmented dataset for querying. 
While it is theoretically possible to generate an infinite number of such noisy variants per data point, we cap the size of the maximum possible augmented dataset $\inferdata$ to $|\protectdata|\cdot(|\protectdata| - 1)$ to make it comparable to \sysS{} below.

\textbf{\sysS{}.} 
\sysR{} introduces unrelated noise that may cause the resulting data to deviate further from the distribution of $\privdata$. \sysS{} is thus proposed as an alternative to better preserve the characteristics of $\privdata$.
After applying the first layer of $\epsilon$-LDP random noise to each data point in $\privdata$ to produce $\protectdata$, we combine pairwise samples of $\protectdata$ for all pairs by averaging corresponding pixel values. Each augmented inference data point \(d_{\text{infer}_i} \in \inferdata\) is computed as 
\(d_{\text{infer}_i} = avg(d_{\text{protected}_i}, d_{\text{protected}_j}) \text{ where }  
i,j = 1, 2, \dots, |\protectdata|, i \neq j\).
This process results in at most $|\protectdata|\cdot(|\protectdata| - 1)$ permutations in $\inferdata$. Since averaging is symmetric,
pairs \((i,j)\) and \((j,i)\) are identical; in our implementation, we retain these
ordered samples to match the experimental query budget of \sysR{}, but deduplication
could reduce the query count without changing the set of unique superimposed images.

The mechanism-level guarantee above also covers structure-aware providers that know the \sysS{} construction. In particular, when sufficiently many pairwise averages are observed, a provider may recover protected records algebraically:
$d_{\text{protected}_i}=\mathrm{avg}(d_{\text{protected}_i},d_{\text{protected}_j})+\mathrm{avg}(d_{\text{protected}_i},d_{\text{protected}_k})-\mathrm{avg}(d_{\text{protected}_j},d_{\text{protected}_k})$.
Such recovery does not violate the formal guarantee: the recovered $d_{\text{protected}_i} \in \protectdata$ is already the output of $\mathcal{A}_{\epsilon}$, and the attack does not recover the private sample $d_{\text{priv}_i}$ or break the $\epsilon$-LDP privacy guarantee.

\section{Evaluation}\label{sec:eval}
In this section, we present the evaluation results and address our research questions in Section~\ref{sec:intro} with empirical analysis.

\paragraph{\textbf{Experimental setup.}}
We run our experiments on two machines. One has two GPUs, NVIDIA GeForce RTX 3090 and 4090, with 24GB of dedicated memory, and an Intel 12th Gen i7-12700 CPU with 12 cores and 64GB of RAM. The other has two NVIDIA GeForce RTX 4090 GPUs and an AMD Ryzen Threadripper PRO 5955WX CPU with 16 cores and 64GB of RAM. The underlying operating systems are 64-bit Ubuntu 22.04.3 LTS and Ubuntu 24.04 LTS, respectively. We use Python 3.9.7 and PyTorch v2.1.2 with CUDA 12.1. 

We evaluate \sys{} on three diverse datasets: SVHN~\cite{svhn}, Fashion-MNIST~\cite{fashionmnist}, and PathMNIST from MedMNIST2D~\cite{medmnist} for medical imaging in pathology. 
For ML models, we use ResNet-152 as ${\teacher}$, and ResNet-18 and MobileNetV2 as ${\student}$.
The $\student$ models are initialized with random weights.

To construct $\privdata$, we train each $\teacher$ on a subset of the original data rather than the default training split: 35k samples for Fashion-MNIST, 48,257 for SVHN, and 89,996 for PathMNIST. The remaining data is divided into a candidate pool for $\privdata$ and a holdout set, $\valdata$, used to evaluate $\student$'s generalizability. The candidate pools contain 25k, 25k, and 10,004 samples for Fashion-MNIST, SVHN, and PathMNIST, respectively, while $\valdata$ contains 10k, 26,032, and 7,180 samples. To model users with limited private data, we set $|\privdata|=1{,}500$ in Section~\ref{subsec:RQ1eval}, sampled uniformly across classes from the candidate pool. We study the effect of varying $|\inferdata|$ and $|\privdata|$ in Section~\ref{subsec:dataset_sensitivity}. All experiments are repeated over three random seeds and subset splits; dependent two-sample t-tests confirm that all reported improvements are statistically significant ($p<0.05$).

\begin{figure}[th!]
    \centering
    \includegraphics[width=0.8\linewidth]{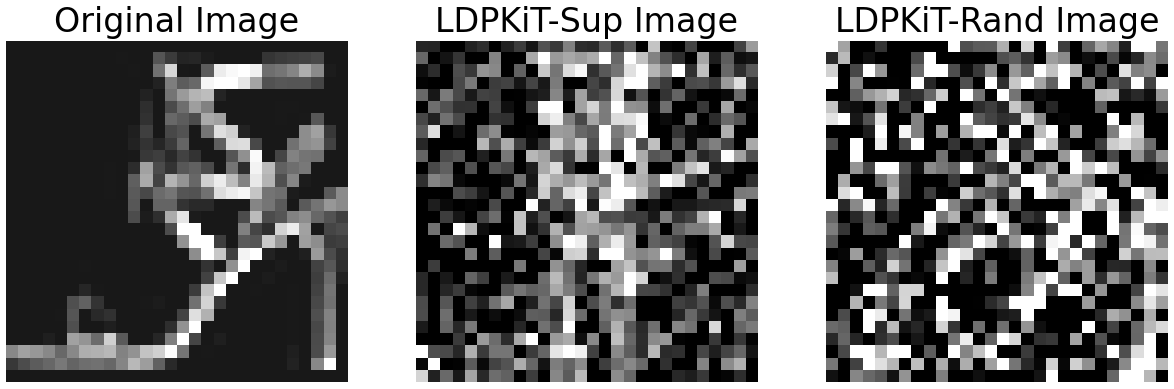}
    \caption{Example of a Fashion-MNIST sample with label 5 (sandal) and $\epsilon=1.5$.}
    \label{fig:fmnist_noised_sample} 
\end{figure} 

We report SIDP accuracy using $\teacher$'s labels on $\protectdata$, while \sys{} accuracy is measured on $\privdata$ using $\student$ trained with $\teacher$'s labels on $\inferdata$. Following industry practice~\cite{apple_white,Orr_2017}, we choose $\epsilon$ values so that SIDP achieves 1.5--3$\times$ random-guessing accuracy, giving $\teacher$ minimal but useful signal while maintaining robustness to reconstruction attacks (Section~\ref{subsec:privacy_analysis}). We use $\epsilon \in {2.0,1.5,1.25}$ for SVHN, ${2.0,1.5}$ for Fashion-MNIST, and ${10.0,7.0}$ for PathMNIST, consistently across SIDP, \sysR{}, and \sysS{}. Figure~\ref{fig:fmnist_noised_sample} shows Fashion-MNIST samples in $\inferdata$ generated by \sysR{} and \sysS{}.

\begin{figure*}[ht!]
    \centering
    \begin{subfigure}[th!]{0.495\textwidth} 
        \centering
        \includegraphics[width=\textwidth]{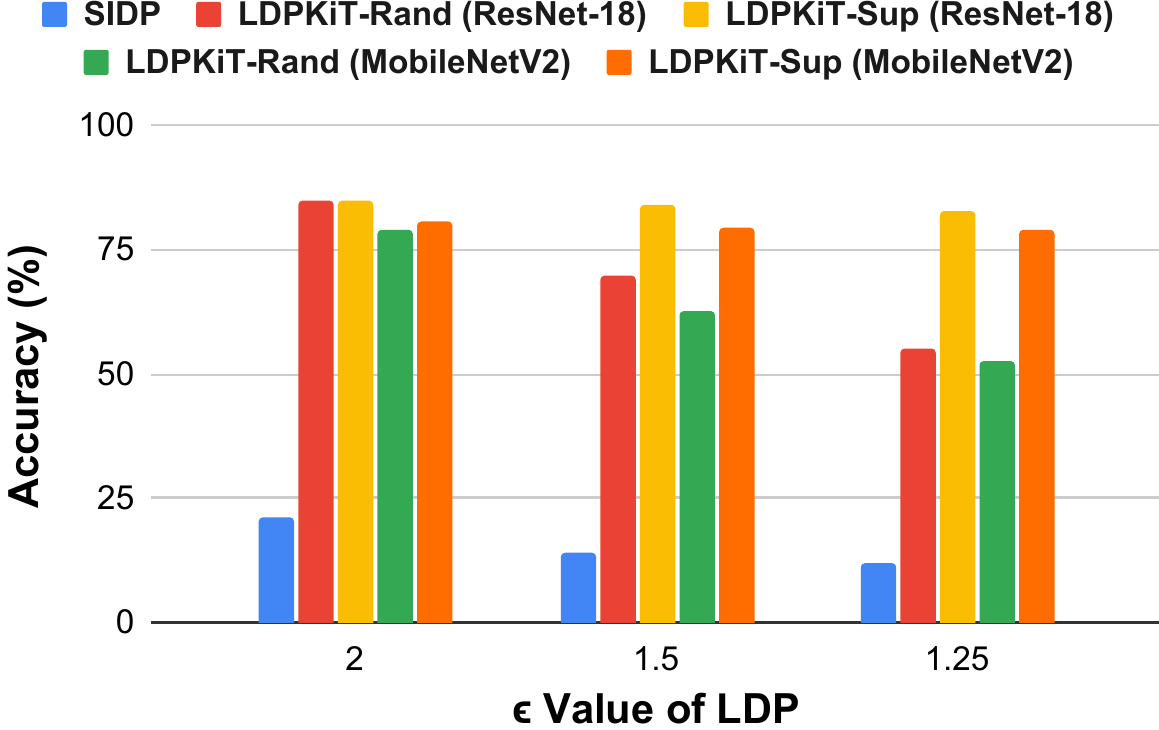}
        \caption{SVHN}
        \label{subfig:svhn_testacc}
    \end{subfigure}
    \hfill
    \begin{subfigure}[th!]{0.495\textwidth}
        \centering
        \includegraphics[width=\textwidth]{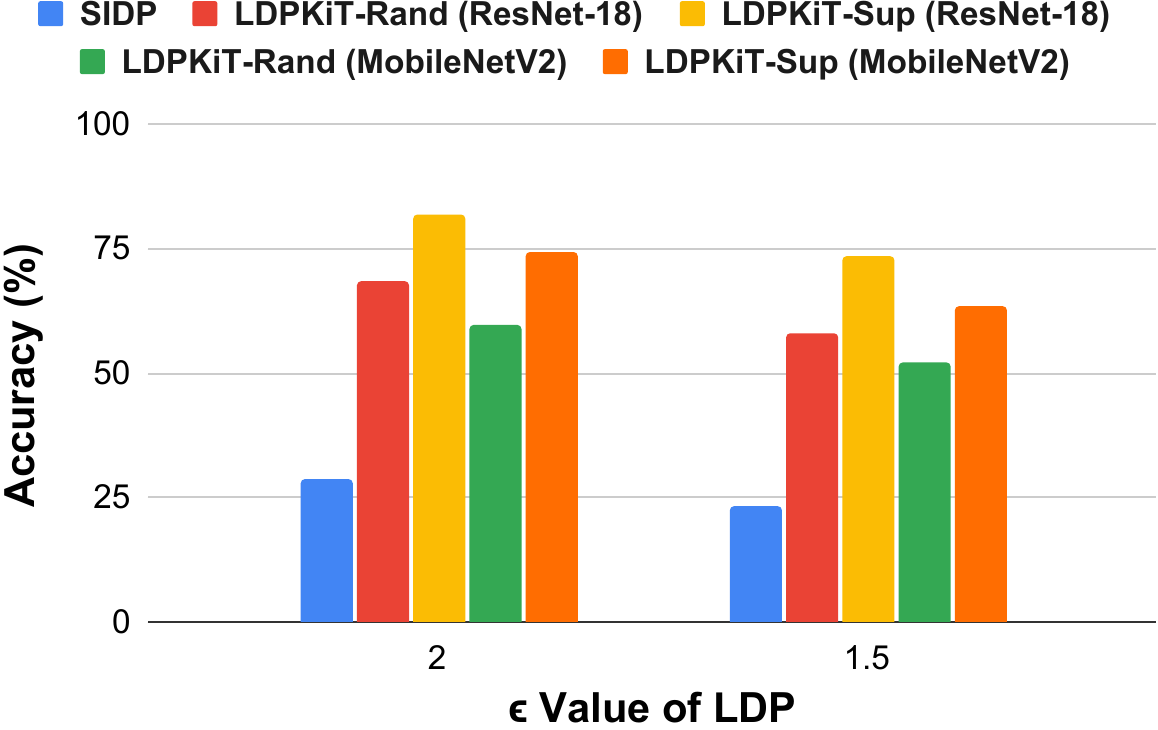}
        \caption{Fashion-MNIST}
        \label{subfig:fmnist_testacc}
    \end{subfigure}
    \caption{Inference accuracy on $\privdata$ for SVHN and Fashion-MNIST under different $\epsilon$ values, comparing SIDP using $\teacher$ with \sysR{} and \sysS{} distillation using $\student$. See Figure~\ref{subfig:pathmnist_testacc} for PathMNIST results.}
    \label{fig:ResNet18_mobilenet_test_accs}
\end{figure*}

\subsection{RQ1: \sys{}'s Utility Recovery on $\privdata$}\label{subsec:RQ1eval}
In this section, we evaluate whether \sys{} improves prediction accuracy on $\privdata$ compared to directly using $\teacher$'s returned labels on $\protectdata$ (\ie SIDP). To simulate a realistic scenario, we set $|\privdata|$ to 1,500, representing a practical amount of private data available to a user. For variable control, we use the full $\inferdata$, of size $|\protectdata|\cdot(|\protectdata| - 1)$ for querying $\teacher$ and training $\student$ to minimize variations from random subset splits. 

Without privacy protection, $\teacher$'s average accuracies on noise-free $\privdata$ of Fashion-MNIST, SVHN, and PathMNIST are 93.33\%, 94.73\%, and 86.2\%, respectively. The accuracies on $\valdata$ are 93.22\%, 96.30\% and 81.24\%, respectively.

To answer RQ1, we compare the last-epoch accuracies of $\student$ on $\privdata$ with $\teacher$'s SIDP accuracies on $\protectdata$ in Figure~\ref{fig:ResNet18_mobilenet_test_accs} and Figure~\ref{subfig:pathmnist_testacc}. Overall, \sys{} consistently improves accuracy over SIDP. For example, when $\epsilon=2.0$ on SVHN, ResNet-152's SIDP accuracy is only 21.07\%, while both \sysR{} and \sysS{} recover ResNet-18 accuracy to about 85\%.
However, \sysR{} provides limited gains in some cases. On Fashion-MNIST, ResNet-18 reaches only 68.52\% and 58.07\% accuracy at $\epsilon=2.0$ and $\epsilon=1.5$, respectively, while MobileNetV2 remains below 60\%. On PathMNIST, ResNet-18 also performs poorly, and MobileNetV2 with \sysR{} can even underperform SIDP. These results suggest that adding arbitrary random noise does not generate sufficiently representative training data.
By contrast, \sysS{} consistently outperforms \sysR{}. Superimposing noised $\privdata$ samples produces more useful $\inferdata$ than applying random noise. For instance, on Fashion-MNIST with $\epsilon=2.0$, \sysS{} raises ResNet-18 accuracy to 81.78\%, compared with 28.69\% for SIDP and 68.52\% for \sysR{}. On PathMNIST with $\epsilon=10.0$, \sysS{} reaches 82.11\%, far above SIDP at 28.74\% and \sysR{} at 43.90\%. On SVHN, despite SIDP accuracies of only 11.87\%--21.07\%, \sysS{} consistently achieves over 80\% accuracy with both ResNet-18 and MobileNetV2.
These results show that \sys{}, especially \sysS{}, effectively recovers much of the utility lost to $\epsilon$-LDP noise while maintaining the same per-sample $\epsilon$-LDP protection.

\begin{figure*}[th!]
    \centering
    \begin{subfigure}[th!]{0.325\textwidth} %
        \centering
        \includegraphics[width=\textwidth]{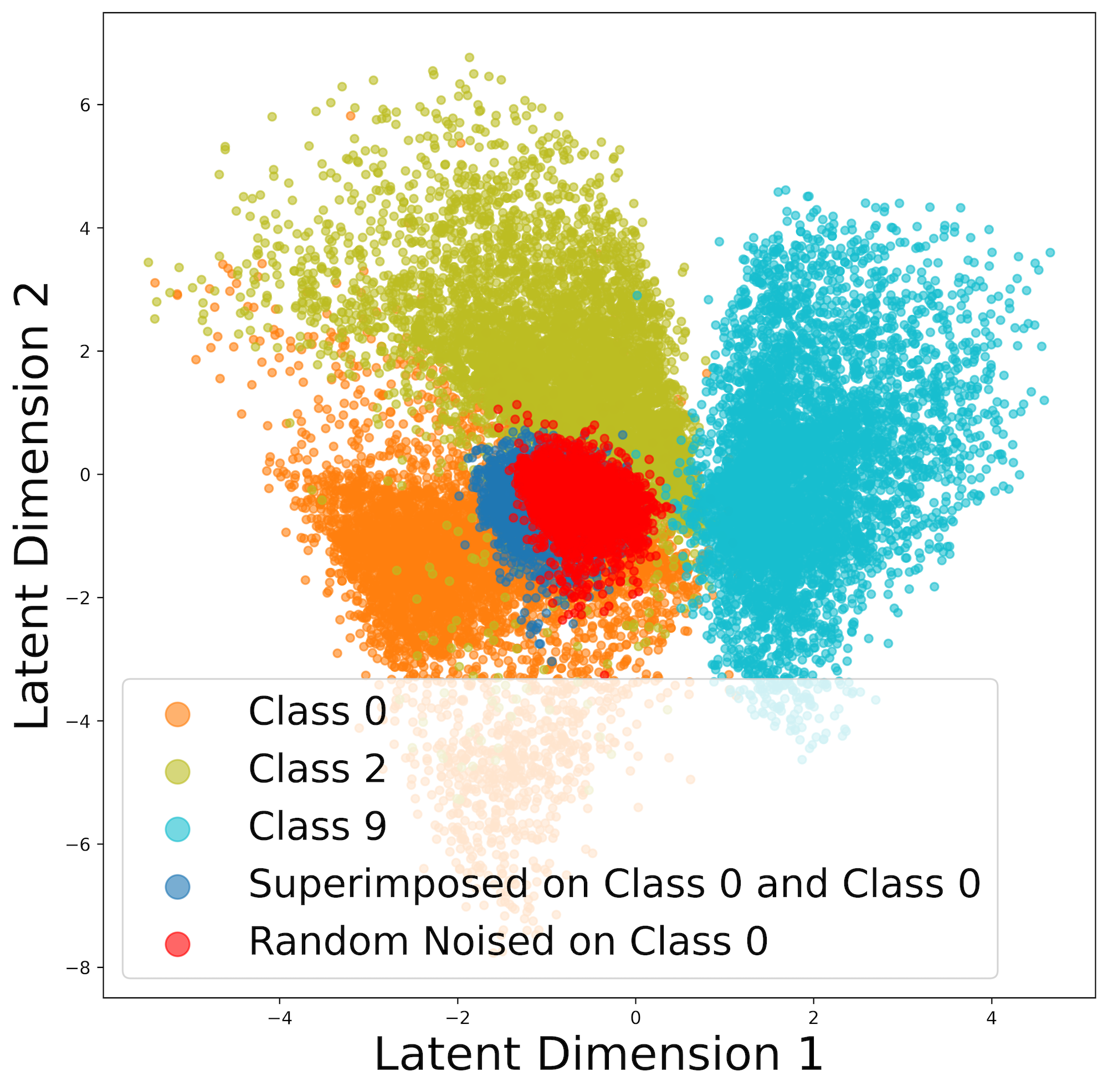}
        \caption{(0,0)}
        \label{subfig:fmnist00}
    \end{subfigure}
    \hfill
    \begin{subfigure}[th!]{0.325\textwidth}
        \centering
        \includegraphics[width=\textwidth]{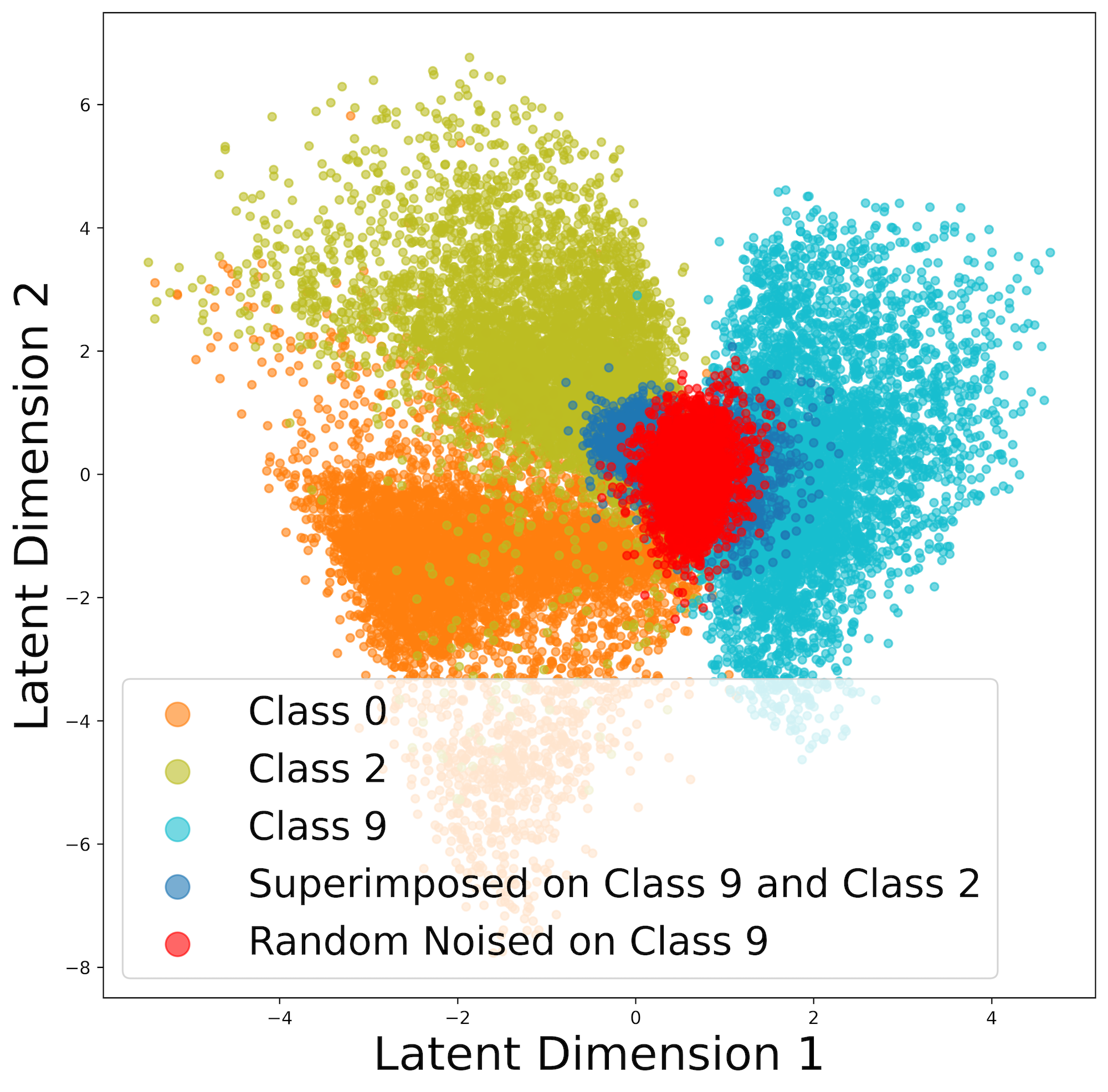}
        \caption{(9,2)}
        \label{subfig:fmnist92}
    \end{subfigure}
    \hfill
    \begin{subfigure}[th!]{0.325\textwidth}
        \centering
        \includegraphics[width=\textwidth]{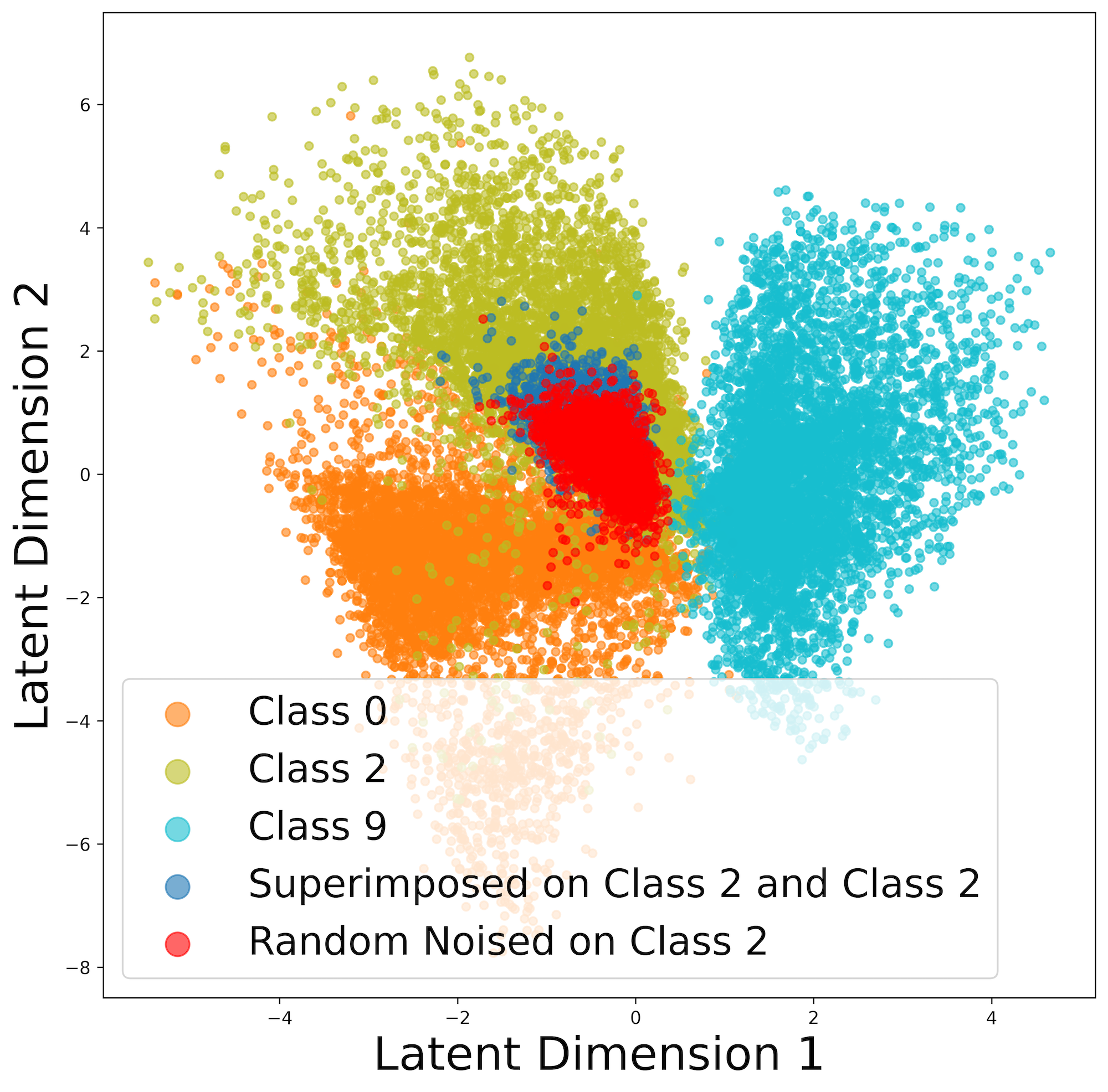}
        \caption{(2,2)}
        \label{subfig:fmnist22}
    \end{subfigure}
    
    \caption{Latent space plots of Fashion-MNIST class triplets (C0-T-shirt/top , C2-pullover, C9-ankle boot) and privacy-protected noisy data clusters generated with \sysR{} and \sysS{} ($\epsilon=2.0$).
    }

    \label{fig:fmnist_latent_029}
\end{figure*}

\begin{table}[ht!]
\caption{Euclidean distances between centroids of clusters in Figure~\ref{fig:fmnist_latent_029}.
$d(C_{\text{N}}, C_X)$ is the Euclidean distance between the Noisy and Class X cluster centroids.
}
\label{tab:fmnist_latent_029_euclidean}

\begin{center}
\scriptsize
\newcolumntype{Y}{>{\centering\arraybackslash}X}
\begin{tabularx}{\linewidth}{lXXXXX}
\toprule
\textbf{Figure} & \textbf{Strategy} & \textbf{Class(es)} & \bm{$d(C_{\textbf{\text{N}}}, C_0)$} &\bm{ $d(C_{\textbf{\text{N}}}, C_2)$} & \bm{$d(C_{\textbf{\text{N}}}, C_9)$} \\
\midrule
\multirow{2}{*}{\ref{subfig:fmnist00}} 
& Sup & (0,0) & \textbf{1.9835} & 2.5191 & 3.2017 \\
& Rand & 0 & \textbf{2.1195}  & 2.4666 & 2.9169  \\
\midrule
\multirow{2}{*}{\ref{subfig:fmnist92}} 
& Sup & (9,2) & 3.1652 & \textbf{2.3538} & \textbf{2.0524}  \\
& Rand & 9 & 3.0937 & 2.5595 & \textbf{1.9333} \\
\midrule
\multirow{2}{*}{\ref{subfig:fmnist22}} 
& Sup & (2,2) & 2.9662 & \textbf{1.7351} & 2.9171\\
& Rand & 2 & 2.7084 & \textbf{1.9433} & 2.7115 \\
\bottomrule

\end{tabularx}

\end{center}

\end{table}

\subsection{RQ2: Latent Space Analysis}\label{subsec:latent}

Section~\ref{subsec:RQ1eval} shows that \sys{} improves accuracy on $\privdata$ over SIDP, with \sysS{} consistently outperforming \sysR{}. We hypothesize that \sysS{} is more effective because superimposition produces augmented samples that better preserve the target-class structure, yielding more informative data for knowledge transfer. To examine this, we analyze the latent space of $\inferdata$ by training a VAE on $\epsilon$-LDP-noised samples from three randomly selected classes, using $\epsilon=2.0$ for Fashion-MNIST and SVHN and $\epsilon=7.0$ for PathMNIST. To obtain separable class clusters, we replace the conventional reconstruction loss with the triplet margin loss. With the trained VAE, we visualize two additional noisy clusters: one generated by \sysS{} from two target classes and one generated by \sysR{} from a single class. We then compute the Euclidean distances between these noisy clusters and the original clusters to assess whether \sysS{} provides a latent-space learning advantage over \sysR{}.

Figure~\ref{fig:fmnist_latent_029} visualizes the Fashion-MNIST latent space for Classes 0, 2, and 9. The \sysS{} clusters are generated by superimposing noisy samples from target-class pairs, whereas the \sysR{} clusters are generated by applying additional random noise to samples from a single class. The plots show that \sysS{} generally produces samples closer to the target-class regions than \sysR{}.
For example, in Figures~\ref{subfig:fmnist00} and~\ref{subfig:fmnist22}, the \sysS{} clusters lie closer to their corresponding target-class clusters than the \sysR{} clusters. Similarly, in Figure~\ref{subfig:fmnist92}, the \sysS{} cluster overlaps more with Classes 2 and 9, while the \sysR{} cluster is less representative, positioned between the clusters of Classes 0, 2, and 9.

Table~\ref{tab:fmnist_latent_029_euclidean} reports Euclidean distances between cluster centroids and supports the visual trends. We note, however, that the Euclidean distance between centroids is not always an ideal measure of similarity between clusters. For example, in Figure~\ref{subfig:fmnist92}, \sysS{} visually overlaps more with Class 9, but because it combines samples from both Classes 2 and 9, its distribution is more dispersed; consequently, the \sysR{} centroid can appear slightly closer to the Class 9 centroid despite being less representative overall.

To better understand the latent-space representations, we compute Kullback-Leibler (KL) divergence~\cite{kullback1951}. Let $C_T$ denote the target triplet cluster, $C_S$ the \sysS{} cluster generated from superimposed target-class pairs, and $C_R$ the \sysR{} cluster generated from noisy samples of the same triplet classes, with $C_S$ and $C_R$ matched in size. We define the \textit{Divergence Ratio (DR)} as
\[
\text{DR}(\bar{C}_R, \bar{C}_S) =
\frac{D_{\text{KL}}(\bar{C}_T \,\|\, \bar{C}_R)}
     {D_{\text{KL}}(\bar{C}_T \,\|\, \bar{C}_S)},
\]
where $\bar{C}_T$, $\bar{C}_S$, and $\bar{C}_R$ are normalized distributions. $\text{DR}>1$ indicates that \sysS{} is closer to the target triplet distribution than \sysR{}.

Across 130 triplets from SVHN, Fashion-MNIST, and PathMNIST, \sysS{} achieves lower KL divergence than \sysR{} in 68.56\% of cases. This supports our hypothesis that \sysS{} more often preserves the target latent distribution, consistent with its stronger accuracy in Section~\ref{subsec:RQ1eval}. However, KL divergence does not fully explain accuracy gains, and a more complete statistical characterization remains future work.

\begin{table*}[t]
\caption{Impact of $|\privdata|$ and $|\inferdata|$, with ResNet-18 ($\student$).}

\label{tab:sensitivity}
\begin{center}
\scriptsize
\setlength{\tabcolsep}{0.5pt}
\newcolumntype{Y}{>{\centering\arraybackslash}X}
\begin{tabularx}{\linewidth}{@{}Yrc|Y|YY|YYY|YYYY|YYYY@{}}
\toprule

\textbf{Dataset} & \bm{$\epsilon$} &  \textbf{\sys{}} & \multicolumn{14}{c}{\textbf{Accuracy on $\privdata$ (\%)}} \\\midrule
& $|\privdata|$ &  & 125 & \multicolumn{2}{c|}{250} & \multicolumn{3}{c|}{500} & \multicolumn{4}{c|}{1k} & \multicolumn{4}{c}{1.5k} \\\cmidrule{4-17}
& $|\inferdata|$ &  & 15.5k & 15.5k & 62.3k & 15.5k & 62.3k & 250k & 15.5k & 62.3k & 250k & 500k & 15.5k & 62.3k & 250k & 500k \\\midrule
\multirow{2}{*}{\makecell{Fashion-\\MNIST}} & \multirow{2}{*}{2.0} & Rand & 14.8 & 18.4 & 36.4 & 17.7 & 32.5 & 52.9 & 18.1 & 34.5 & 57.9 & 63.4 & 14.8 & 34.7 & 57.2 & 64.7  \\
 & & Sup & 25.9 & 28.9 & 31.4 & 24.2 & 37.6 & 53.0 & 25.1 & 47.2 & 66.5 & 73.7 & 25.1 & 45.1 & 69.0 & 74.4 \\ \midrule
\multirow{2}{*}{SVHN} & \multirow{2}{*}{1.5} & Rand & 8.8 & 11.4 & 6.9  & 9.4 & 6.9  & 34.1 & 10.5 & 9.9  & 35.3 & 51.3 & 10.4 & 10.0 & 32.8 & 48.6 \\
 & & Sup & 9.1 & 9.7 & 12.7 & 10.1 & 15.3 & 45.7 & 10.2 & 20.0 & 60.1 & 69.1 & 10.1 & 14.3 & 62.1 & 76.9 \\ \midrule
\multirow{2}{*}{\makecell{Path-\\MNIST}} & \multirow{2}{*}{10} & Rand & 24.4 & 24.8 & 29.9 & 22.9 & 32.8 & 44.4 & 24.5 & 30.8 & 50.4 & 43.4 & 23.6 & 29.0 & 43.7 & 54.4 \\
 & & Sup & 41.1 & 41.4 & 47.5 & 44.0 & 55.0 & 69.9 & 40.9 & 48.5 & 73.3 & 77.0 & 43.0 & 57.2 & 74.1 & 76.8  \\

\bottomrule
\end{tabularx}
\end{center}
\end{table*}

\begin{figure*}[t]
    \centering
    \begin{subfigure}[th!]{0.495\textwidth}
        \centering
        \includegraphics[width=\textwidth]{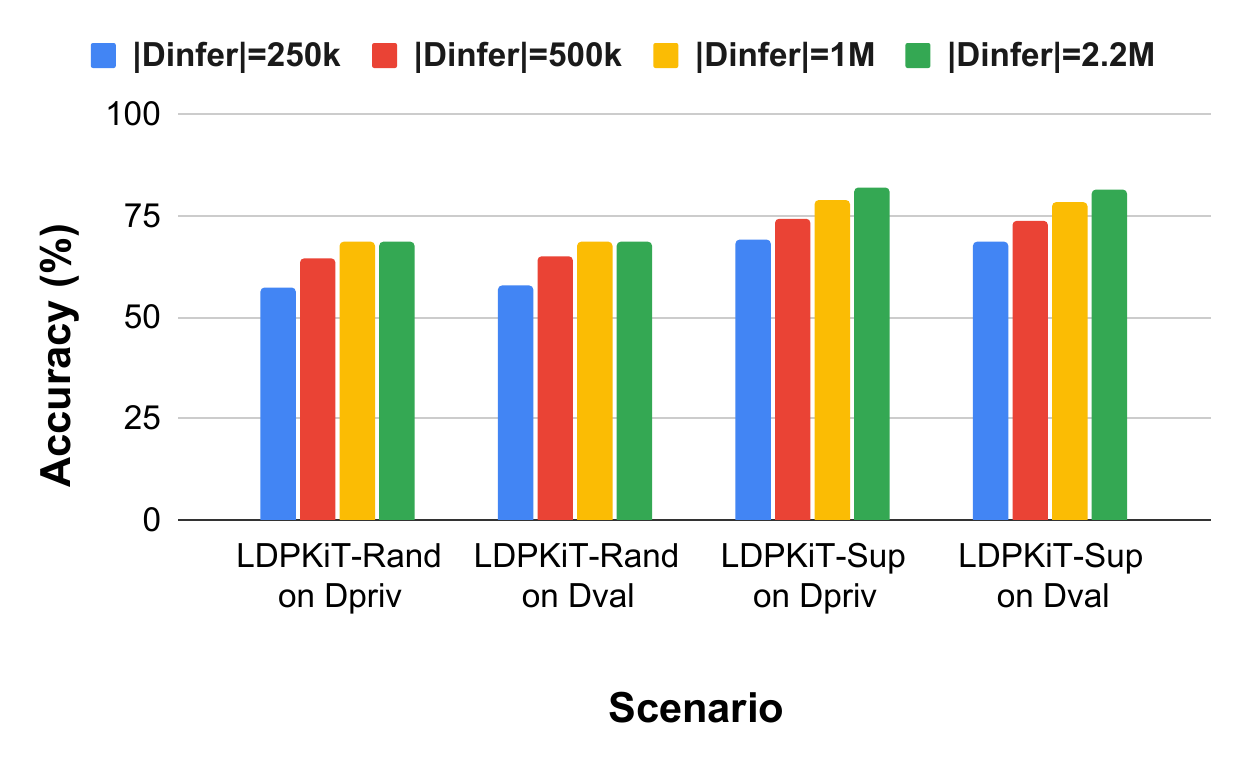}
        \caption{Fashion-MNIST ($\epsilon = 2.0$)}
        \label{subfig:fmnist_infer_sensitivity_2}
    \end{subfigure}
    \hfill
    \begin{subfigure}[th!]{0.495\textwidth}
        \centering
        \includegraphics[width=\textwidth]{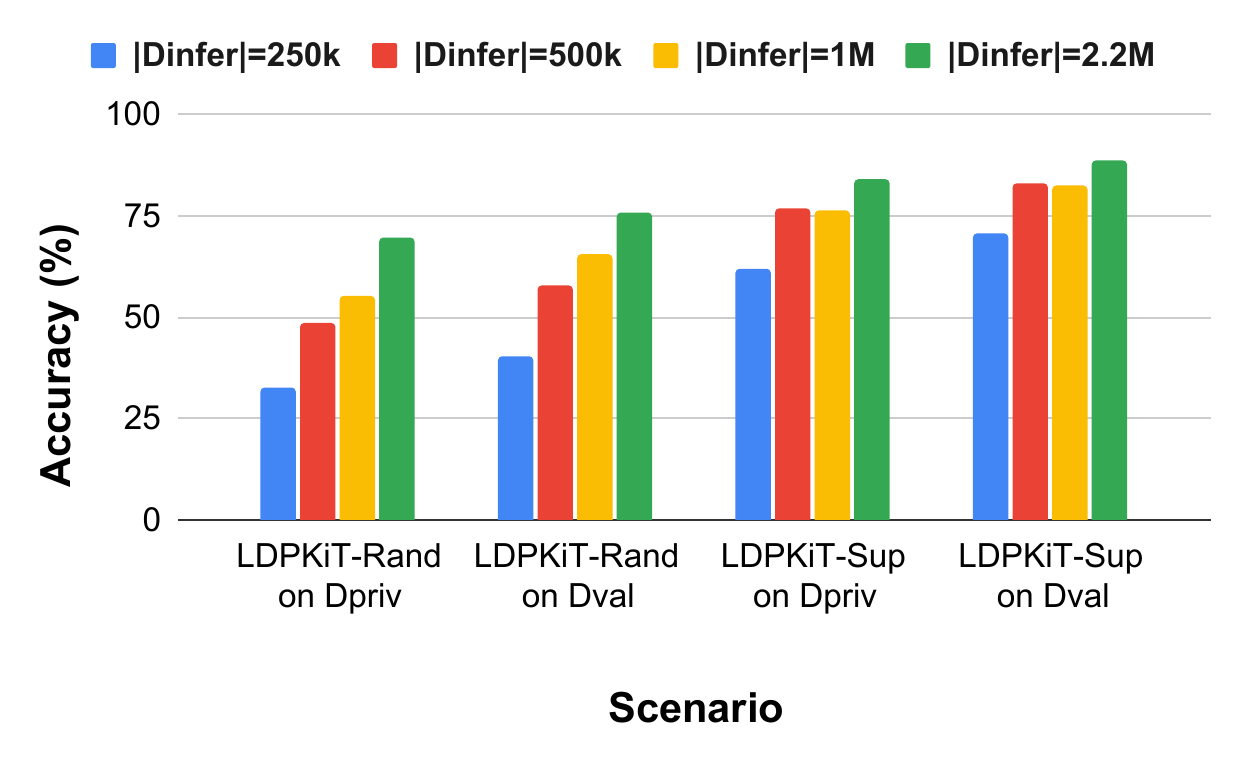}
        \caption{SVHN ($\epsilon = 1.5$)}
        \label{subfig:svhn_infer_sensitivity_1.5}
    \end{subfigure}
    \caption{Impact of $|\inferdata|$ on ResNet-18 $\student$ accuracy for Fashion-MNIST and SVHN, with $|\privdata|=1.5k$. See Figure~\ref{subfig:pathmnist_infer_sensitivity_10} for PathMNIST results.}
    \label{fig:infer_sensitivity}
\end{figure*}

\begin{figure*}[hbpt!]
    \centering
    \begin{subfigure}[ht]{0.43\textwidth}
        \centering
        \includegraphics[width=\textwidth]{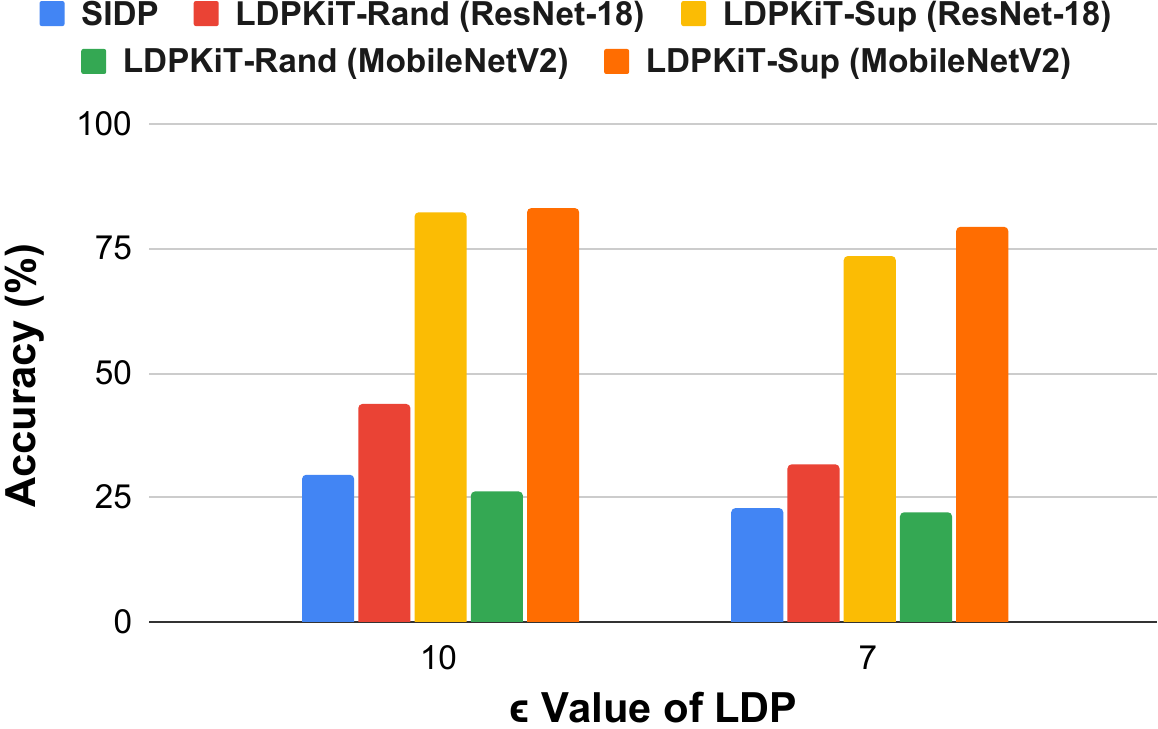}
        \caption{Inference accuracy}
        \label{subfig:pathmnist_testacc}
    \end{subfigure}
    \hfill
    \begin{subfigure}[ht]{0.5\textwidth}
        \centering
        \includegraphics[width=\textwidth]{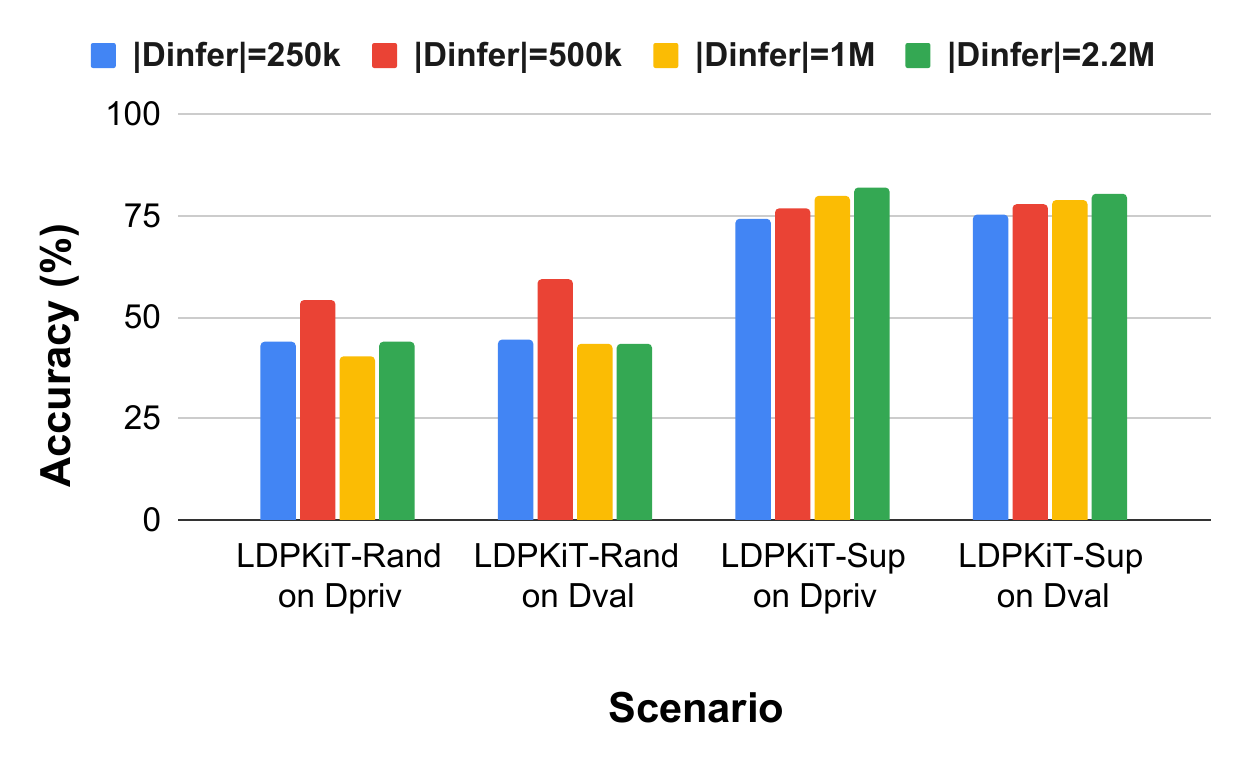}
        \caption{Impact of $|\inferdata|$ ($\epsilon = 10.0$)}
        \label{subfig:pathmnist_infer_sensitivity_10}
    \end{subfigure}
    \caption{PathMNIST results: inference accuracy and impact of $|\inferdata|$, evaluated with ResNet-18 $\student$ and $|\privdata|=1.5k$. Complemantary to Figures~\ref{fig:ResNet18_mobilenet_test_accs} and \ref{fig:infer_sensitivity}.}
    \label{fig:pathmnist_results}
\end{figure*}

\subsection{RQ3: Impact of $|\inferdata|$ and $|\privdata|$} \label{subsec:dataset_sensitivity}

We analyze the trade-off among privacy exposure ($|\privdata|$), query cost ($|\inferdata|$), and accuracy by varying the number of private samples used for augmentation and the number of generated queries used to train $\student$. As shown in Table~\ref{tab:sensitivity}, we construct multiple $\inferdata$ datasets from $\privdata$ at different scales. For example, when $|\privdata|=125$, the maximum $|\inferdata|$ is 15,500, computed as $|\protectdata|\cdot(|\protectdata|-1)$. When $|\privdata|=250$, $|\inferdata|$ can either remain 15,500 through random selection or increase to a maximum of 62,250. To isolate the effect of $|\privdata|$, we fix $|\inferdata|$ and compare different $|\privdata|$ values; to isolate the effect of $|\inferdata|$, we fix $|\privdata|$ and vary $|\inferdata|$. We exhaustively evaluate these settings for $|\privdata|$ from 125 to 1.5k and report ResNet-18 ($\student$) accuracies.

The results show that \sys{} is relatively insensitive to $|\privdata|$, indicating that users can achieve similar utility with fewer private samples and therefore lower privacy exposure. For example, on Fashion-MNIST with $\epsilon=2.0$, when $|\inferdata|=15{,}500$, the maximum accuracy gap across different $|\privdata|$ settings is below 5\%; when $|\inferdata|=500k$, the gap is only about 0.69\%. Thus, with a reasonable $|\inferdata|$, users can generate effective queries from a smaller $\privdata$. One exception occurs when $|\inferdata|=250k$, where increasing $|\privdata|$ improves accuracy, likely because more private samples increase the chance of generating representative and diverse training data.

In contrast, $|\inferdata|$ has a stronger impact on accuracy because it directly determines the amount of training data and query cost. Although $|\privdata|$ constrains the maximum feasible $|\inferdata|$, sufficient $\inferdata$ is crucial for effective training. For example, on PathMNIST with $\epsilon=10.0$ and $|\privdata|=1.5k$, \sysR{} improves from 23.56\% accuracy at $|\inferdata|=15{,}500$ to 54.39\% at $|\inferdata|=500k$, while \sysS{} improves from 42.99\% to 76.79\%. Conversely, on SVHN, $\student$ remains near 10\% accuracy when $|\inferdata|=15{,}500$, regardless of $|\privdata|$.

Figure~\ref{fig:infer_sensitivity} and Figure~\ref{subfig:pathmnist_infer_sensitivity_10} further illustrate the effect of $|\inferdata|$ under a fixed $|\privdata|=1.5k$. \sysS{} requires a sufficiently large $\inferdata$ to achieve adequate accuracy, such as 250k for PathMNIST and 500k for Fashion-MNIST and SVHN. Increasing $|\inferdata|$ improves accuracy, but the gains diminish beyond 500k. For instance, on Fashion-MNIST with $\epsilon=2.0$, ResNet-18 achieves 74.35\%, 78.88\%, and 81.78\% accuracy when $|\inferdata|$ is 500k, 1M, and 2.2M, respectively. Thus, querying all possible $\inferdata$ points is unnecessary; \sysS{} can achieve satisfactory accuracy using only about 20\% of $\inferdata$. 

To estimate the API query cost, we use OpenAI's listed GPT-5 nano pricing as of May 2026 (\$0.05 per 1M input tokens and \$0.40 per 1M output tokens)~\cite{openai_gpt5nano_pricing}
and OpenAI's image-token accounting rules for patch-based image inputs (\(2.46\) input tokens per image)~\cite{openai_image_vision_pricing}. Assuming a one-token class-label output, each query costs about \(5.23 \times 10^{-7}\) USD, so \(2.2\)M queries cost about \$1.15 under a lower-bound estimate, or \$3.35 with a minimal 20-token text prompt. Rate limits, latency, batching, and API policies remain practical constraints.

Finally, the $\inferdata$ subsets used in these experiments are randomly selected, and there is currently no universal lower bound on $|\privdata|$ that guarantees reliable performance because the required size depends on the task, $\teacher$ complexity, LDP noise level, and the diversity of $\privdata$. In practice, users can perform small-scale pilot runs by gradually increasing $|\inferdata|$ and monitoring accuracy trends, while systematic estimation of practical $|\privdata|$ thresholds remains future work.

\begin{table}[t]
\centering
\scriptsize
\caption{Classification accuracy and image similarity under various $\epsilon$ settings. Similarity is measured against the original images before and after denoising.
}
\newcolumntype{Y}{>{\centering\arraybackslash}X}

\begin{tabularx}{\linewidth}{@{}lcYYYYY@{}}
\toprule
\textbf{Dataset} & $\boldsymbol{\epsilon}$ & \textbf{Recon.} & \makecell{\textbf{Acc. (\%)}}$\uparrow$ & \textbf{MSE} $\downarrow$ & \textbf{SSIM} $\uparrow$ & \textbf{LPIPS} $\downarrow$ \\
\midrule
\multirow{6}{*}{SVHN} 
  & \multirow{2}{*}{2.0}  & Before           & 16.80 & 0.0681 & 0.1210 & 0.7591 \\
  &                       & After            & 14.20 & 0.1524 & 0.0825 & 0.7183 \\
  & \multirow{2}{*}{1.5}  & Before           & 15.30 & 0.0929 & 0.0850 & 0.7747 \\
  &                       & After            & 14.40 & 0.1436 & 0.0829 & 0.7301 \\
  & \multirow{2}{*}{1.25} & Before           & 12.50 & 0.1098 & 0.0681 & 0.7826 \\
  &                       & After            & 13.00 & 0.1382 & 0.0810 & 0.7347 \\
\midrule
\multirow{4}{*}{\makecell{Fashion-\\MNIST}} 
  & \multirow{2}{*}{2.0}  & Before           & 16.10 & 0.0942 & 0.3869 & 0.5568 \\
  &                       & After            & 16.70 & 0.1257 & 0.3180 & 0.4492 \\
  & \multirow{2}{*}{1.5}  & Before           & 14.80 & 0.1123 & 0.3406 & 0.5928 \\
  &                       & After            & 16.30 & 0.1395 & 0.2953 & 0.5044 \\
\midrule
\multirow{4}{*}{\makecell{Path-\\MNIST}} 
  & \multirow{2}{*}{10.0} & Before           & 22.78 & 0.0132 & 0.4823 & 0.2237 \\
  &                       & After            & 23.89 & 0.0853 & 0.3870 & 0.2593 \\
  & \multirow{2}{*}{7.0}  & Before           & 17.56 & 0.0155 & 0.3936 & 0.2507 \\
  &                       & After            & 21.89 & 0.1012 & 0.3110 & 0.2805 \\
\bottomrule
\end{tabularx}

\label{tab:full_denoising_metrics}
\end{table}

\subsection{Evaluation of Data Reconstruction Risks }\label{subsec:privacy_analysis}
To rigorously evaluate whether the amount of LDP noise added to each private image is sufficient, we audit the privacy guarantees of \sys{} following the methodology of~\cite{schwethelm2025visualprivacyauditingdiffusion}. Specifically, we conduct data reconstruction attacks using a DDPM trained to model the image prior (\ie trained on the original images). Our experiment spans three datasets: SVHN, Fashion-MNIST, and PathMNIST. For each dataset, we randomly select a subset of 1,000 images (900 for PathMNIST) processed with \sysS{}, and quantitatively measure reconstruction success with standard image-similarity metrics: MSE, VGG-based LPIPS, and SSIM. We report results in Table~\ref{tab:full_denoising_metrics}. Our measurements show that the reconstructed images are generally more dissimilar from the original images after reconstruction (\ie lower SSIM and higher MSE). We did not observe successful reconstructions under our evaluated attack setting. We also show that a fully trained classifier (ResNet-152) shows near-random accuracy on the reconstructed images.

\section{Related work}\label{sec:rw}
\paragraph{\textbf{Knowledge transfer techniques.}} Knowledge distillation is a knowledge transfer technique that distills a large teacher model into a smaller student model while preserving model performance~\cite{hinton2015distilling,KDsurvey}.
The conventional use case is model compression, enabling deployment in a resource-restricted environment. Knowledge can be transferred in different forms~\cite{KDsurvey}, and such techniques can also be used adversarially.
Model extraction is an adversarial setting in which the attacker reproduces a model stealthily by stealing its parameters, decision boundaries, or functionalities. 
It demonstrates that an iterative query-based knowledge transfer process from a high-performance model can be performed via a prediction query interface~\cite{tramer2016stealing, zhangideal, hetal}. 
Model extraction can be challenging without knowledge of the victim model's training data distribution~\cite{truong2021data}. Successful model extractions with partial or zero knowledge of the victim model and training data~\cite{practicalbbattack,truong2021data} may require more information than hard labels.
Related defenses~\cite{deepmarks,dziedzic2022increasing,PoT} are in active research.
Rather than model compression or adversarial extraction for the purpose of theft, we adopt query-based knowledge transfer techniques in a non-adversarial setting to recover the utility loss introduced by LDP noise while protecting users' private data.

\paragraph{\textbf{Noise injection and differential privacy (DP).}}
DP can be applied either globally~\cite{papernotPATE,Abadi_2016} or locally~\cite{RAPPOR}, both offering provable privacy guarantees. Global DP relies on a trusted curator who accesses raw sensitive data before adding noise to aggregate statistics. To avoid this trust assumption, \sys{} adopts local DP, where the user perturbs each query before sending it to the ML service. This gives users direct control over privacy protection, but typically reduces utility compared with global DP at the same privacy level~\cite{privacy_at_scale}.
Unlike training-time methods such as DP-SGD~\cite{Abadi_2016}, \sys{} protects inputs at inference time. Related split-computation schemes inject noise into intermediate DNN representations~\cite{notjustprivacy}, but they assume white-box access to a partitioned model.

\paragraph{\textbf{Federated Learning (FL).}}
FL has emerged as a popular approach to address privacy concerns by enabling collaborative model training without directly sharing users' raw data with a central server~\cite{li2023privacy,truong2021privacy}. 
However, FL primarily protects users' training data, which is an orthogonal concern to our setting, and it does not mitigate privacy risks during inference. Moreover, FL is vulnerable to gradient inversion attacks that reconstruct local training data from gradient updates, which necessitates extra privacy protection~\cite{zhang2022surveygradientinversionattacks}.
Finally, FL assumes that users possess labelled local datasets, which is not applicable in our case, where the objective is to annotate unlabelled data.
Concerns regarding potential data reconstruction attacks and the integrity of the server imply that relying solely on using shared models from service providers like FL is insufficiently secure and privacy-preserving; therefore, FL is not an ideal solution in our scenario. 

\paragraph{\textbf{Others.}} Another approach is encrypted inference using homomorphic encryption, but such methods incur high computational overhead~\cite{gazelle}. On SVHN, \sys{} becomes more efficient once the query set reaches roughly 800 samples: full knowledge transfer on 800 samples takes 20.73 seconds, compared with 22.984 seconds for a single Lancelot inference~\cite{lancelot} and 484.184 seconds for OpenFHE~\cite{OpenFHE}. Moreover, \sys{} performs knowledge transfer once, after which inference on data from the same distribution runs locally with negligible overhead. However, encryption-based schemes continue to incur remote per-query costs, leading to higher cumulative overhead for labelling larger datasets.
The alternative is hardware-assisted inference using Trusted Execution Environments (TEEs). Slalom protects remote inference by executing computation inside a TEE~\cite{tramer2019slalom}. However, TEEs still access the original data and remain vulnerable to side-channel attacks~\cite{ lipp2020meltdown}. \sys{} avoids this threat model because it never transmits raw data; any privacy leakage is instead bounded by the LDP mechanism.

\section{Limitations}\label{sec:discussion}

Our per-record $\epsilon$-LDP guarantee is distribution-free and does not rely on an \iid assumption. However, correlations among records may still allow inferences about related records or users and can weaken privacy due to mutual information leakage, so the semantic privacy interpretation is strongest when samples are approximately independent.
In practice, many applications use de-duplication to maintain or approximate independence. For example, workflows built on Amazon S3 Inventory, Athena, AWS Lambda, and S3 Batch Operations can identify and remove duplicate objects to reduce storage costs~\cite{Lim_2024}. Similarly, generative AI pipelines commonly de-duplicate LLM training data to improve efficiency and performance while reducing privacy leakage from redundant content~\cite{maini2024llmdatasetinferencedid,carlini2021extractingtrainingdatalarge,lee2022deduplicatingtrainingdatamakes}.

Currently, \sys{} only supports the image modality, where superimposition intuitively translates to pixel-level operations. Although textual data does not have a direct counterpart to superimposition, several conceptual approaches can be explored. These include embedding-level averaging, in which two textual embeddings are blended; token-level mixing, which involves interleaving tokens from two samples; and semantic-level merging, such as combining two sentences into a coherent paraphrased summary. Each interpretation may require a distinct adaptation of the LDP scheme. We leave the exploration of these textual superimposition methods as a direction for future research.

We also extended the evaluation of \sys{} to the ImageNet1K dataset.
For reference, the official torchvision ResNet-18 model reports a top-1 accuracy of 69.758\% on ImageNet1K~\cite{torchvision_resnet18_2025}. In our evaluation, a randomly selected subset of 15 classes was used. \sysS{} with ResNet-18 ($\student$) achieved 63\% top-1 accuracy with 15,000 $\privdata$ samples at $\epsilon=5$, whereas the baseline SIDP attained only 19\% accuracy under identical conditions. Further enhancing the performance of \sys{} on larger datasets is left for future work.

\section{Conclusion}\label{sec:conclusion}

\sys{} introduces a privacy-preserving inference framework for protecting sensitive user data when interacting with remote ML model providers. By injecting LDP noise into inference queries, \sys{} protects user privacy even when queries are logged, retained, or exposed through later infrastructure compromise.
The key contribution of \sys{} is its integration of LDP with a two-layer augmentation mechanism, \sysR{} and \sysS{}, which leverages LDP's post-processing property to improve utility without weakening privacy. The first layer provides baseline LDP protection, while the second layer generates an augmented dataset for local model distillation. In particular, \sysS{} uses superimposition-based augmentation to produce samples that better approximate the private data distribution than \sysR{}'s random-noise strategy.
Evaluation results show that \sysS{} effectively recovers prediction accuracy while preserving privacy, with larger gains under stronger noise levels corresponding to stronger privacy guarantees. Overall, \sys{} offers a privacy-preserving design aligned with responsible data-use and privacy-by-design principles.

\begin{credits}
\subsubsection{\ackname}
Support for this research was provided in part by DND-IDEaS Contract MN3-01, NSERC Discovery Grant RGPIN-2026-07548 and NSERC-CSE Grant ALLRP 588144-23. David Lie is supported in part by Tier 1 Canada Research Chair CRC-2019-00242 and Kexin Li was supported by an SRI Fellowship. Researchers funded through the NSERC-CSE Research Communities Grants do not represent the Communications Security Establishment Canada or the Government of Canada. 
Any part research, opinions or positions they produce as part of this initiative do not represent the official views of the Government of Canada.

\subsubsection{\discintname}  %
The authors have no competing interests to declare.
\end{credits}

\bibliographystyle{splncs04}
\bibliography{ref}

\end{document}